\documentclass[lettersize,journal]{IEEEtran}
\usepackage{amsmath,amsfonts}
\usepackage{algorithmic}
\usepackage{array}
\usepackage[caption=false,font=footnotesize,labelfont=sf,textfont=sf]{subfig}
\usepackage{textcomp}
\usepackage{stfloats}
\usepackage{url}
\usepackage{verbatim}
\usepackage{graphicx}
\usepackage{subfig}

\usepackage{tabularx}        
\usepackage{booktabs}        
\usepackage{multirow}        
\usepackage{array}           
\usepackage{makecell}        
\usepackage{threeparttable}  
\usepackage{lineno}
\usepackage{tcolorbox}

\def\BibTeX{{\rm B\kern-.05em{\sc i\kern-.025em b}\kern-.08em
    T\kern-.1667em\lower.7ex\hbox{E}\kern-.125emX}}
\usepackage{balance}
\begin{document}
\title{Unsupervised Mutual Learning of Discourse Parsing and \\ Topic Segmentation in Dialogue}
\author{Jiahui Xu, Feng Jiang, Anningzhe Gao, Luis Fernando D'Haro,~\IEEEmembership{Member,~IEEE} and Haizhou Li,~\IEEEmembership{Fellow,~IEEE}}

\markboth{Journal of Transactions on Audio, Speech, and Language Processing,~Vol.~18, No.~9, February~2025}{Xu \MakeLowercase{\textit{et al.}}: Unsupervised Mutual Learning of Discourse Parsing and \\ Topic Segmentation in Dialogue}

\maketitle

\begin{abstract}
In dialogue systems, discourse plays a crucial role in managing conversational focus and coordinating interactions. It consists of two key structures: rhetorical structure and topic structure. The former captures the logical flow of conversations, while the latter detects transitions between topics. Together, they improve the ability of a dialogue system to track conversation dynamics and generate contextually relevant high-quality responses. These structures are typically identified through discourse parsing and topic segmentation, respectively. However, existing supervised methods rely on costly manual annotations, while unsupervised methods often focus on a single task, overlooking the deep linguistic interplay between rhetorical and topic structures.
To address these issues, we first introduce a unified representation that integrates rhetorical and topic structures, ensuring semantic consistency between them. Under the unified representation, we further propose two linguistically grounded hypotheses based on discourse theories: (1) Local Discourse Coupling, where rhetorical cues dynamically enhance topic-aware information flow, and (2) Global Topology Constraint, where topic structure patterns probabilistically constrain rhetorical relation distributions. Building on the unified representation and two hypotheses, we propose an unsupervised mutual learning framework (UMLF) that jointly models rhetorical and topic structures, allowing them to mutually reinforce each other without requiring additional annotations. We evaluate our approach on two rhetorical datasets (STAC and Molweni) and three topic segmentation datasets (DialSeg711, Doc2Dial, and TIAGE). Experimental results demonstrate that our method surpasses all strong baselines built on pre-trained language models (PLMs) and achieves competitive performance to state-of-the-art large language models (LLMs) such as GPT-4o. Furthermore, when applied to LLMs (Qwen2-1.5B-Instruct and Llama3-8B-Instruct), our framework achieves notable improvements, demonstrating its effectiveness in improving discourse structure modeling, even when integrated with powerful models.
\end{abstract}


\section{Introduction}

\begin{figure}[htbp]
\centering
\includegraphics[width=\linewidth]{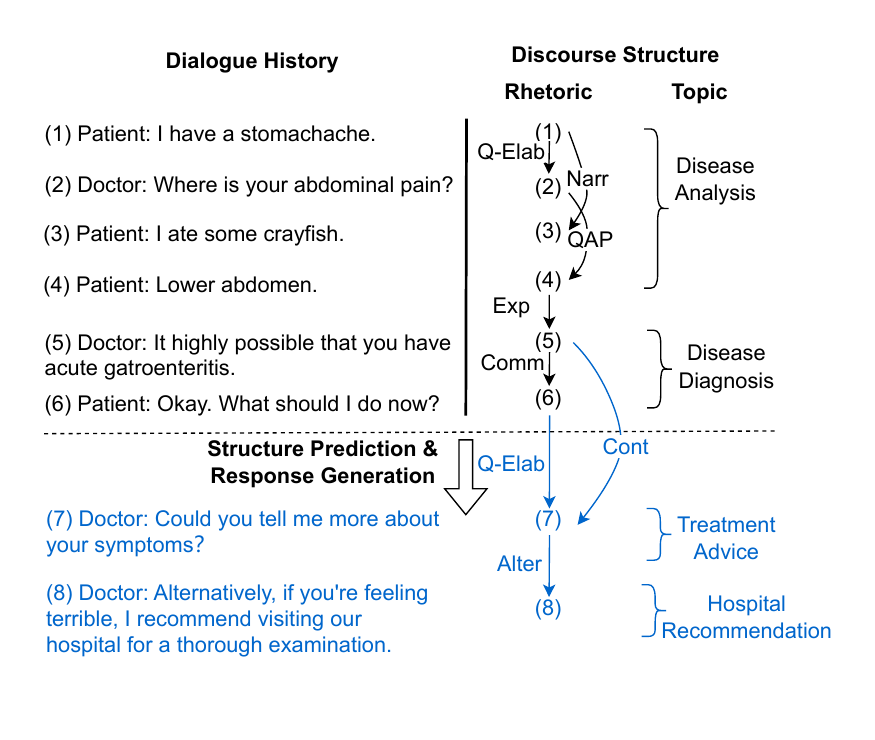} 
\caption{An Illustration of Rhetorical and Topic Structures in Medical Consulting. (Q-Elab: Question-Elaboration, Narr: Narration, QAP: Question-Answer-Pair, Exp: Explanation, Cont: Continuation, Alter: Alternation, Comm: Comment.)}
\label{fig:image1}
\end{figure}

\IEEEPARstart{D}{iscourse} serves as a framework for communication in dialogue systems by organizing the flow of information and articulating intentions. It helps participants achieve their communicative goals, manage conversational focus, and coordinate interactions \cite{van1995discourse}. In discourse theory, it can be deconstructed into a tripartite cognitive framework: Linguistic Structure, Intentional Structure, and Attention State \cite{grosz1986attention}. The latter two correspond respectively to rhetorical structure~\cite{asher2003logics} and topic structure~\cite{todd2003topics} in contemporary discourse analysis. Rhetorical structure captures the logical relationships between discourse units, defining how utterances cohere through rhetorical relations~\cite{asher2003logics}. Topic structure governs topic transitions, determining whether the ongoing topic should be maintained or shifted~\cite{todd2003topics, chafe1994discourse}. Together, these structures enable dialogue systems to effectively manage conversational progression and generate coherent, contextually appropriate responses~\cite{grosz1986attention, stede2022discourse}. Parsing these two structures is essential for balancing multi-party interactions and ensuring smooth, user-centered responses, particularly in complex scenarios such as negotiations~\cite{putnam2010negotiation} and medical consultations~\cite{liu2024context}. A concrete example is illustrated in Fig.~\ref{fig:image1}.

Rhetorical and topic structures are typically detected through discourse parsing~\cite{marcu2000theory} and topic segmentation~\cite{hearst1997text}, respectively. Discourse parsing identifies rhetorical relations between discourse units~\cite{asher2003logics, asher2012reference}, such as Question-Answer Pairs (QAPs) and Comments, providing a high-level representation of dialogue pragmatics. This process usually involves two steps: (1) extracting plain structures without discourse relations and (2) annotating discourse relations. In this work, we focus on the first and fundamental step, as it has been empirically shown to benefit dialogue generation~\cite{sohn2023tod} and various downstream tasks~\cite{louis2010discourse, xu2019discourse, xiao2020we, jia2020multi}.

Conversely, topic segmentation groups utterances into coherent topic clusters, enabling the system to maintain or shift conversational focus appropriately~\cite{todd2011analyzing}. It is often framed as a binary classification task, identifying whether a topic boundary exists between two adjacent utterances. Such segmentation plays a crucial role in high-level dialogue management~\cite{xu2021topic, fu2023delving, chernyavskiy2024groundhog} and contributes to generating meaningful, context-aware responses~\cite{yin2023ctrlstruct}.

\paragraph{Challenges and Limitations}
Building upon established discourse corpora~\cite{asher2016discourse, li2020molweni, feng2020doc2dial, xu2021topic}, current supervised methods leverage pre-trained language models (PLMs) for discourse parsing~\cite{he2021multi, chi2022structured} and topic segmentation~\cite{lin2023topic}. However, due to the high cost of manual annotations, recent works have shifted towards unsupervised methods~\cite{xing2021improving, li2023discourse, gao2023unsupervised} for general discourse processing.

Despite these advances, existing approaches treat discourse parsing and topic segmentation as independent tasks, overlooking their linguistic interdependencies. However, discourse theory rigorously establishes their connection: Rhetorical relations are influenced by discourse intentions, which are often encoded within topic structures~\cite{grosz1986attention}; Topic structure formation relies on fine-grained discourse relations to establish coherence~\cite{stede2022discourse}. Although some studies attempt to bridge these two tasks by leveraging one to assist the other~\cite{jiang2021hierarchical, huber2022predicting, xing2022improving}, these unidirectional methods still rely on annotated data and suffer from cascading errors in multi-step pipelines. As a result, there remains a fundamental gap between their linguistic relationship and their disjoint modeling.

\paragraph{Our Approach}
To bridge this gap, we introduce a novel unified representation and propose two linguistically grounded hypotheses based on discourse theory~\cite{asher2003logics, todd2011analyzing, van1983strategies, mann1988rhetorical}, asserting that rhetorical and topic structures mutually enhance each other. Based on this, we propose an unsupervised mutual learning framework (UMLF)~\cite{zhang2018deep} to jointly model discourse parsing and topic segmentation without requiring additional annotations. 

Specifically, our framework first introduces a unified representation that integrates rhetorical and topic structures, ensuring semantic consistency and enabling mutual learning, as illustrated in Fig.~\ref{fig: relation}. Then, we propose two hypotheses from two perspectives to integrate these two structures: (1) Local Discourse Coupling: Rhetorical structure cues dynamically enhance topic-aware information flow; (2) Global Topology Constraint: Topic structure patterns probabilistically constrain rhetorical relation distributions.

For implementation, we first extend previous representations of rhetorical and topic structures~\cite{li2023discourse, gao2023unsupervised} into two separate matrices. We then design local and global aggregators to capture cross-structural interactions, corresponding to our hypotheses. Finally, we introduce a differentiable loss function to align and integrate both structures within a unified representation, enforcing semantic consistency. Through this mutual learning process, we can obtain simultaneous discourse parsing and topic segmentation from the common structure under the unified representation.

Our main contributions are threefold:

    (1) We introduce a unified representation of rhetorical and topic structures, demonstrating their semantic consistency, and propose two linguistically grounded hypotheses for their mutual enhancement.
    
    (2) We propose an unsupervised mutual learning framework for joint discourse parsing and topic segmentation without the need for additional annotations.
    
    (3) We conduct extensive evaluations on five benchmark datasets, demonstrating the robustness and effectiveness of our approach across both million-scale PLMs and billion-scale LLMs.

\section{Related Work}

\subsection{Unsupervised Topic Segmentation}
In a dialogue, the topic refers to a collection of permanent events, conditions, and references that align with the speaker's partially active discourse~\cite{van1983strategies, chafe1994discourse}, which can indicate the stage of the conversation and the current focus of the discussion~\cite{todd2003topics, todd2011analyzing}. Topic segmentation aims to obtain topic boundaries by measuring the topic similarity between adjacent utterances. Most methods involve two parts for unsupervised topic segmentation in dialogue~\cite{van1983strategies, chafe1994discourse}: a statistical technique or deep neural model is employed to assess the topic similarity between the two sides of each potential topic boundary, followed by a segmentation algorithm such as TextTiling~\cite{hearst1997text} to identify the segment boundaries. Early studies~\cite{choi2000advances, glavavs2016unsupervised, riedl2012topictiling} typically assess topic similarity through dialogue coherence or semantic similarity, often based on surface features such as lexical overlap. Later, Song et al.~\cite{song2016dialogue} assessed semantic similarity using pre-trained word embeddings. This method was later extended to use sentence embeddings from pre-trained language models~\cite{solbiati2021unsupervised, he2022space}. Xing et al.~\cite{xing2021improving} further proposed the Coherence Scoring Model (CSM), which employs utterance-pair coherence to assess topic similarity. More recently, Gao et al.~\cite{gao2023unsupervised} proposed an unsupervised Dialogue Topic Segmentation framework (DialSTART), which learns topic-aware utterance representations from unlabeled dialogue data through neighboring utterance matching and pseudo-segmentation.

\subsection{Unsupervised Discourse Parsing}
Based on the Rhetorical Structure Theory (RST)~\cite{mann1988rhetorical} for documents, the Segmented Discourse Representation Theory (SDRT)~\cite{asher2003logics, asher2012reference} is a popular rhetorical structure representation of dialogue. It models rhetorical relations using a tree structure, where arcs between discourse units capture their functional relationships. Discourse parsing aims to uncover the internal structure of a multi-participant conversation by finding all the discourse links. For unsupervised discourse parsing in dialogue~\cite{asher2003logics, asher2012reference}, previous work has primarily focused on developing various decoding algorithms~\cite{muller2012constrained, li2014text, afantenos2012modelling, perret2016integer} to construct the discourse tree. To address the issue of data sparsity, several studies have explored approaches such as weak supervision and transfer learning~\cite{badene2019data, liu2021improving}. After that, Li et al.~\cite{li2023discourse} leveraged Pre-trained Language Models (PLMs), finetuning them with other dialogue datasets to learn relevant structural information and utilizing the model's encoder attention outputs to generate a rhetorical structure matrix.

\begin{figure}[!ht]
    \centering
    \includegraphics[width=1.0\linewidth]{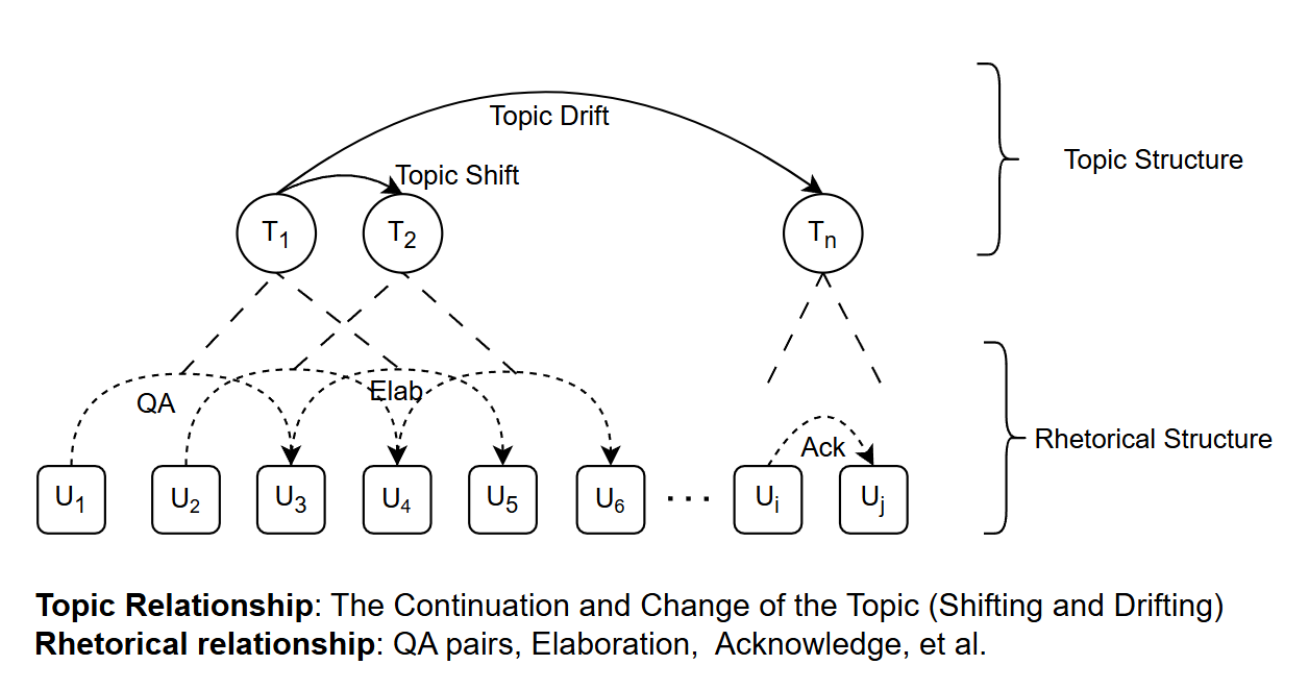}
    \caption{The unified representation of rhetorical and topic structures for dialogues.}
    \label{fig: two_relations}
\end{figure}

\subsection{Unidirectional Learning between Rhetorical and Topic Structures}
Since both rhetorical and topic structures are considered higher-level discourse structures~\cite{grosz1986attention, stede2022discourse, louis2012coherence, muangkammuen2020neural}, some efforts have explored using one to assist the other for better performance. For example, Xing et al.~\cite{xing2022improving} introduced a neural topic segmentation model that integrates rhetorical structures into the underlying topic segmentation framework, leveraging topic coherence for more accurate topic boundary predictions. Conversely, Jiang et al.~\cite{jiang2021hierarchical} proposed MDParserTS, a Macro Discourse Parser that uses Topic Segmentation mechanisms to leverage implicit topic boundaries, thereby enhancing the construction of hierarchical macro rhetorical structures. However, both approaches still rely on supervised training using annotated data from prior tasks, where rhetorical structure aids topic segmentation or vice versa, but their influence remains unidirectional, lacking mutual reinforcement.

\section{Mutual Learning for Discourse Parsing and Topic Segmentation}

The semantic relations between dialogue segments encompass both topic-related connections, such as topic shift and topic drift~\cite{lin2023topic}, which reflect shifts in the focus of the conversation, and rhetorical connections, such as QA pairs and elaborations~\cite{asher2016discourse}, which convey the underlying intentions and functions of the ongoing discourse~\cite{xie2021tiage}. In light of this, we propose a unified representation of these two structures for dialogue, as illustrated in Fig.~\ref{fig: two_relations}. 
Building on this unified framework, we explore the linguistic interactions and relationships between these two structures and correspondingly present two hypotheses: one in which rhetorical structure supports topic structure and another where topic structure influences rhetorical structure. 
Based on these two hypotheses, we design an unsupervised mutual learning framework to jointly model the common structure through semantic consistency between them.~\footnote{We will release our code at \url{https://github.com/Jeff-Sue/URT}.}

\subsection{The Linguistic Connection between Rhetorical Structure and Topic Structure}
\begin{figure}[ht!]
\centering
\includegraphics[width=1.0\linewidth]{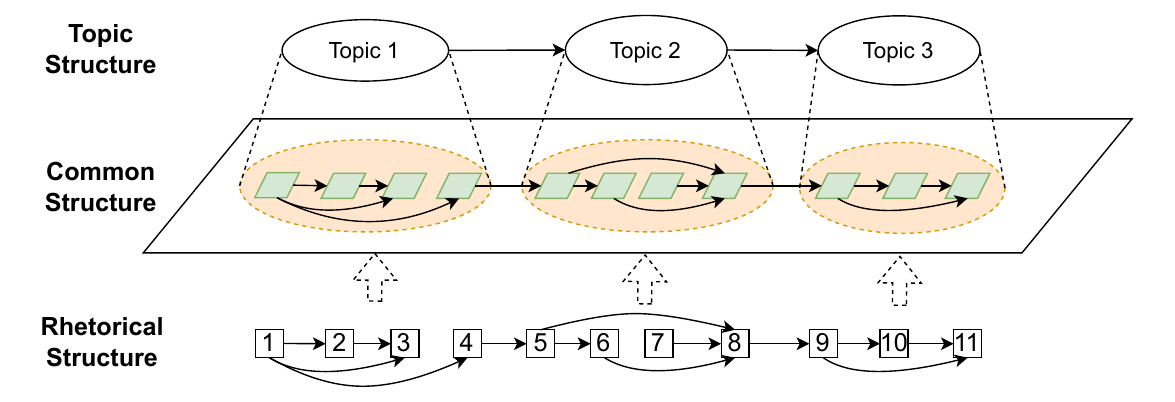}
\caption{An example of rhetorical structure and topic structure promoting each other with 11 utterances. The common structure simultaneously integrates both rhetorical and topic structures while ensuring consistency throughout.}
\label{fig: relation}
\end{figure}

In discourse processing, since rhetorical structure and topic structure differ in representation and analytical approaches, existing works hardly model them simultaneously under a unified framework. However, inherent linguistic connections exist between them under our unified representation, which can mutually reinforce each other, as shown in Fig.~\ref{fig: relation}.

On the one hand, rhetorical structure aligns closely with adjacent (local) text coherence and is strongly correlated with thematic consistency between utterances~\cite{louis2012coherence, muangkammuen2020neural}. Prior work demonstrates that the local rhetorical relationship between phrases or perspectives plays a crucial role in maintaining a well-structured topic~\cite{asher2003logics}. For instance, a cause-effect relation between two utterances often implies a continuity in topic structure. Accordingly, we propose the first hypothesis:

\textbf{Hypothesis 1:} Local Discourse Coupling: From a local perspective, the distribution of rhetorical structures in adjacent utterance blocks dynamically enhances topic-aware information flow, reinforcing rhetorical coherence within topics while reducing segmentation errors across topics.

Similarly, topic structure also supports rhetorical structures by logically dividing extended text passages into smaller segments that are unified by their exploration of a common topic~\cite{stede2022discourse}. Van et al.~\cite{van1995discourse} also argue that discourse topics are intrinsic features of the logical form of coherent discourse or rhetorical relations. Additionally, Stede et al.~\cite{stede2012discourse} highlight that, beyond genre-based structures, topic-based segmentation can effectively partition lengthy discourse into smaller, manageable segments. Accordingly, we propose the second hypothesis:

\textbf{Hypothesis 2:} Global Topology Constraint: From a global perspective, topic segmentation patterns probabilistically constrain the skeleton of the rhetorical structure, reinforcing rhetorical connections within the same topic while minimizing errors across different topics.

Different from prior work in unidirectional learning between rhetorical and topic structures~\cite{jiang2021hierarchical, xing2022improving}, we further propose an unsupervised mutual learning framework (UMLF) to model their bidirectional connections based on our two hypotheses, leveraging semantic consistency for mutual learning.

\begin{figure*}[ht]
    \centering
    \includegraphics[width=0.9\textwidth]{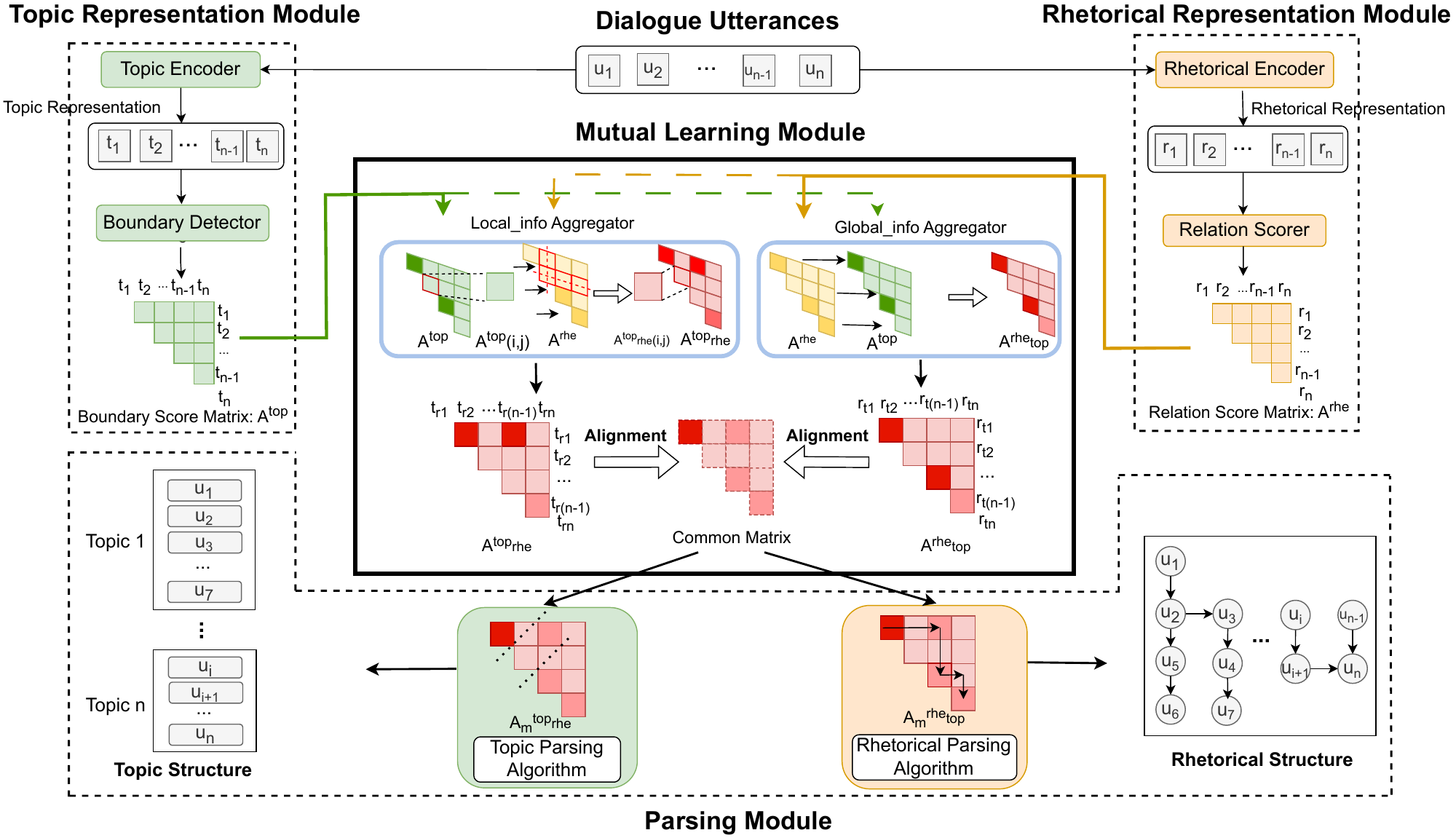}
    \caption{The overall architecture of our unsupervised mutual learning framework. 1) Rhetorical and Topic Representation Module: encode the topic and rhetorical structure information of dialogue utterances into two score matrices. 2) Mutual Learning Module: two aggregators separately integrate local rhetorical information and global topic information into the other structure matrix and align these two fused matrices with a common matrix under the unified representation. 3) Parsing Module: Obtain the topic structure and rhetorical structure by two parsing algorithms.}
    \label{fig: architecture}
\end{figure*}

\subsection{Unsupervised Mutual Learning Framework}

As illustrated in Fig.~\ref{fig: architecture}, the unsupervised mutual learning framework (UMLF) consists of three parts: the Rhetorical and Topic Representation Module, the Mutual Learning Module, and the Parsing Module. Unlike previous work~\cite{jiang2021hierarchical,xing2022improving}, which uses one task to assist another, our unsupervised mutual learning framework allows the two tasks to support each other, eliminating the need for additional annotations of rhetorical and topic structures while simultaneously improving performance in both topic segmentation and discourse parsing.

\subsubsection{Rhetorical and Topic Representation Module}

Given a dialogue containing $n$ utterances, we first obtain their rhetorical and topic semantic representations and extend their scoring matrices using a boundary detector and a relation scorer.

Specifically, we adopt a two-step method~\cite{li2023discourse} to encode the rhetorical structure information into a matrix. First, we utilize the BART Large encoder~\cite{lewis-etal-2020-bart} as the rhetorical encoder. We input the dialogue $D$ containing $n$ utterances, and the resulting [SEP] representation of each utterance from the encoder serves as the rhetorical representations, $H_r = \{r_1, r_2, ..., r_n\}$. Then, we utilize the encoder attention in BART Large as the relation scorer to encode the $n$-dimensional rhetorical representations into an upper-triangular relation score matrix $A^{rhe} \in \mathbb{R}^{n^2}$, where $A^{rhe}_{i,j}$ represents the rhetorical relation score between utterances $u_i$ and $u_j$.

Inspired by Gao et al.~\cite{gao2023unsupervised}'s work on detecting topic boundaries based on coherence and consistency, we encode topic structure information using a topic encoder that incorporates SimCSE~\cite{gao2021simcse} for consistency representation and Next Sentence Prediction BERT~\cite{devlin-etal-2019-bert} for coherence representation. This encoder transforms the dialogue $D$ containing $n$ utterances into topic representations, \textbf{$H_t = \{t_1, t_2, ..., t_n\}$}. Next, we employ a boundary detector to compute the consistency score between any two utterances based on their cosine similarity, while the coherence score is obtained by feeding their coherence representations into a CLS layer. The sum of these two scores is then used as the boundary score between the two utterances. Unlike previous work~\cite{gao2023unsupervised}, which calculates only adjacent utterance pairs, to explore the topic relation from a broader perspective and align with the format of the rhetorical matrix, we obtain an upper-triangular boundary score matrix $A^{top} \in \mathbb{R}^{n^2}$ by applying this process to any pair of utterances, where $A^{top}_{i,j}$ represents the topic boundary score between utterances $u_i$ and $u_j$.

\subsubsection{Mutual Learning Module}

Based on the definition of the unified representation and two linguistic hypotheses, we design a mutual learning module that comprises two key components: the local information aggregator and the global information aggregator. These components integrate rhetorical and topic information, generating two fused matrices with similar structural characteristics. Finally, the module aligns the two fused matrices towards a common structure under the unified representation.

\textbf{Local Information Aggregator.} As outlined in Hypothesis 1 (Local Discourse Coupling), the distribution of rhetorical structures dynamically enhances topic-aware information flow, highlighting the differences between different topics while maintaining similarity within the same topic. To capture this dynamic, we first calculate the relevant topic semantic information flow \(W^{re}\) for each element \(A^{top}_{ij}\) (which represents the topic boundary between utterances \(i\) and \(j\)), as shown in Eq. (\ref{equation1}). Here, \(W_{col}^{top}\) and \(W_{row}^{top}\) are learnable matrices for the columns and rows, respectively, designed to adjust the influence of semantic information from the topic matrix, thereby shaping the semantic flow in accordance with rhetorical structures.

\begin{equation}\label{equation1}
W^{re}_{ij} =  \sum_{k=i}^{n}W_{col}^{top}A^{top}_{kj}+ \sum_{k=1}^{j}A^{top}_{ik}W_{row}^{top}
\end{equation}

Next, to incorporate rhetorical influence into the semantic flow, we perform an element-wise multiplication between the computed topic information matrix \(W^{re}\) and the rhetorical matrix \(A^{rhe}\). This operation results in a local rhetorical matrix \(W^{R}\), which captures the combined influence of both topic semantics and rhetorical structure, as shown in Eq. (\ref{equation2}).

\begin{equation}\label{equation2}
W^{R} = W^{re} \cdot A^{rhe}
\end{equation}

Finally, to obtain the rhetoric-influenced topic matrix, we compute the cross-product of the local rhetorical matrix \(W^{R}\) with the topic matrix \(A^{top}\). This interaction enables the local rhetorical influences to interact with the broader topic semantics, resulting in a rhetoric-enhanced topic matrix \(A^{top_{rhe}}\), which encapsulates both inter-topic distinctions and intra-topic coherence, as shown in Eq. (\ref{equation3}).

\begin{equation}\label{equation3}
A^{top_{rhe}}_{xy} = \sum_{k=1}^{n}W^{R}_{xk}A^{top}_{ky}
\end{equation}

\textbf{Global Information Aggregator.} As described in Hypothesis 2 (Global Topology Constraint), topic segmentation patterns probabilistically constrain the skeleton of the rhetorical structure. Integrating global topic information into the rhetorical structure helps reinforce rhetorical connections within the same topic and minimize errors across different topics. To achieve this, as shown in Eq. (\ref{equation4}), we first modulate the influence of the topic matrix \(A^{top}\) by multiplying it on the left and right with two learnable matrices, \(W_{left}^{top}\) and \(W_{right}^{top}\), which control the extent to which the topic structure influences the rhetorical matrix. We then combine this transformed topic matrix with the original rhetorical matrix \(A^{rhe}\), resulting in a topic-assisted rhetorical matrix \(A^{rhe_{top}}\), which maps the distributional characteristics of the topic matrix onto the rhetorical matrix, thereby achieving our proposed integration.

\begin{equation}\label{equation4}
A^{rhe_{top}} = W_{left}^{top}A^{top}W_{right}^{top} + A^{rhe}
\end{equation}

\textbf{Mutual Alignment.} We align the two fused matrices closer to the common matrix by minimizing the gap between them using Mean Squared Error (MSE) loss, as shown in Eq. (\ref{equation5}). Additionally, we introduce two penalty terms, \(P_1\) and \(P_2\), as shown in Eqs. (\ref{EQ 6}) and (\ref{EQ 7}), to enhance the structural characteristics of the output structure, specifically to prevent the output matrix from being dominated by values close to the median. These penalty terms regulate the variance and mean of the matrix values, ensuring structural differentiation across elements in each matrix.

\begin{equation}\label{EQ 6}
P_1 = std(A_{m}^{top_{rhe}}) + std(A_{m}^{rhe_{top}})
\end{equation}

\begin{equation}\label{EQ 7}
P_2 = \frac{1}{mean(A_{m}^{top_{rhe}})} + \frac{1}{mean(A_{m}^{rhe_{top}})}
\end{equation}

\begin{equation}\label{equation5}
\mathcal{L} = \text{MSE}(A^{top_{rhe}}, A^{rhe_{top}}) - P_1 - P_2
\end{equation}

\subsubsection{Parsing Module}
After alignment and generating the common matrix, we apply TextTiling~\cite{hearst1997text} and Eisner~\cite{eisner1996three} as decoding algorithms for topic segmentation and discourse parsing, respectively, to obtain topic and rhetorical structures simultaneously. The TextTiling algorithm is a classical technique for subdividing texts into multi-paragraph units representing passages or subtopics. It computes the cosine similarity between bag-of-words vectors of sentence gaps and inserts a boundary when the boundary score exceeds a threshold, indicating a lack of coherence between sentences. The Eisner algorithm, based on dynamic programming, continuously merges analysis results of adjacent substrings until the entire sentence is parsed, with parsing items having heads at the ends rather than in the middle. It parses the dialogue into directed rhetorical relation pairs, which are then organized into a tree structure. We chose these two classical algorithms following previous unsupervised work~\cite{li2023discourse, gao2023unsupervised}, ensuring a fair comparison.

\section{Experiment}

To comprehensively validate the effectiveness and robustness of our framework, we conduct the following experiments on five popular topic segmentation and discourse parsing datasets:
(1) For PLM-based models (million-scale parameters), we benchmark against mainstream unsupervised PLM baselines and zero-shot LLMs (billion-scale parameters) requiring no supplemental annotations.
(2) For LLM-based implementations (billion-scale parameters), we perform systematic comparisons of Qwen2-1.5B-Instruct and Llama3-8B-Instruct outputs through: Raw text generation analysis, Parsed results without mutual learning modules, and Our framework.


\subsection{Experiment Settings}

\textbf{Datasets.} Since there is no dataset containing both labeled rhetorical and topic structure, we utilize five diverse datasets with different data sources, types, and construction methods to show the robustness of the method. For discourse parsing, we employ \textbf{STAC}\footnote{\url{https://github.com/chijames/structured_dialogue_discourse_parsing/tree/master/data/stac}}~\cite{asher2016discourse}: a corpus of multi-party chats from an online game, and \textbf{Molweni}\footnote{\url{https://github.com/chijames/structured_dialogue_discourse_parsing/tree/master/data/molweni}}~\cite{li2020molweni}: a machine reading comprehension dataset built over multiparty dialogue. For topic segmentation, we use \textbf{Doc2Dial}\footnote{\url{https://github.com/AlibabaResearch/DAMO-ConvAI/tree/main/dial-start}}~\cite{feng2020doc2dial}: a goal-oriented dialogue dataset grounded in documents from social welfare websites, \textbf{TIAGE}\footnote{\url{https://github.com/HuiyuanXie/tiage}}~\cite{xie2021tiage}: an augmented version of PersonaChat, which is an open-domain dialogue dataset, and \textbf{DialSeg711}~\cite{eric2017key, budzianowski2018multiwoz}: a dataset created by combining dialogues from the MultiWOZ Corpus2 and Stanford Dialogue Dataset, with each dialogue focusing on a single topic. More details of these five datasets are summarized in Table~\ref{tab:dataset_splits_by_type}.

\begin{table}[!ht]
    \centering
    \caption{The Details of Experimental Datasets. DP and TS of task separately represents Discourse Parsing and Topic Segmentation. Utter. means average utterance per dialogue, Rel. means average rhetorical relation per dialogue, and Shif. means the average number of topic shift turns per dialog.}
    \label{tab:dataset_splits_by_type}
    \resizebox{\linewidth}{!}{
    \begin{tabular}{cccccc}
    \toprule
    \textbf{Task} & \textbf{Name} & \textbf{Utte.} & \textbf{Rel.} & \textbf{Shif.}& \textbf{Train/Val/Test} \\
    \hline
     DP & Molweni & 8.8 & 7.8 & - &  8771/883/100 \\
    DP & STAC  & 10. & 11.4 & - & 965/-/116 \\
    \hline
        TS & Doc2Dial & 12.7 & - & 2.9 & 2895/621/621 \\
    TS & TIAGE & 14.8 & - & 3.5 &300/100/100 \\
        TS & DialSeg711 & 27.2 & - & 5.6 &711/-/711 \\
    \bottomrule
    \end{tabular}
    }
\end{table}

\textbf{Evaluation Metrics.} To evaluate discourse parsing, we use the F1 score, calculated based on the predicted arc spans and the corresponding golden arc spans~\cite{chi2022structured}. For topic segmentation, we adopt the $P_k$ error score~\cite{beeferman1999statistical} and Window Difference (WD)~\cite{pevzner2002critique}, which are widely used to assess the performance of segmentation models by measuring the overlap between the ground-truth segments and the model's predictions within a sliding window of a certain size.

\begin{table*}[ht]
\centering
\caption{Performance comparison between ours and baselines in the popular datasets of topic segmentation and discourse parsing. Numbers annotated with Bold and underlined are the best and the second-best results in Millions-Parameter PLMs. The numbers annotated with bold and italics represent the best results among all models.}
\label{main}
\resizebox{\linewidth}{!}{
\begin{tabular}{ccccccccccc}
\toprule
  \multirow{3}{*}{\textbf{Model Size}}& \multirow{3}{*}{\textbf{Method}} & 
  \multicolumn{2}{c}{\textbf{Discourse Parsing}} & \multicolumn{6}{c}{\textbf{Topic Segmentation}} & \multirow{3}{*}{\textbf{Mean}}  \\ \cmidrule(r){3-4} \cmidrule(l){5-10}
 & &  \multicolumn{1}{c}{\textbf{STAC}} &\multicolumn{1}{c}{\textbf{Molweni}} & \multicolumn{2}{c}{\textbf{Doc2Dial}} & \multicolumn{2}{c}{\textbf{TIAGE}} & \multicolumn{2}{c}{\textbf{DialSeg711}} \\
      & & $F1$  & $F1$ & $1-P_k$ & $1-WD$ & $1-P_k$ & $1-WD$ & $1-P_k $ & $1-WD$  \\
\midrule
\multirow{12}{*}{\textit{\begin{tabular}[c]{@{}c@{}}Millions-\\ Parameter PLMs\end{tabular}}} 
&\multicolumn{1}{l}{TeT+CLS(110M)} &- &- & 45.66 & 42.08 & 52.73 & 48.62 & 59.51 & 56.86 & - \\
& \multicolumn{1}{l}{CSM(110M)} & - & - & 54.70 &  50.16 & 52.81 & 49.12 & 75.70 & 73.65 & - \\
& \multicolumn{1}{l}{DialSTART(220M)}   & - & - & 59.95 & 56.12 & 55.66 & 50.24 & 82.14 & 80.20 & - \\
&\multicolumn{1}{l}{PLM4DiscStruct(406M)} & 42.90 & 48.92 & -& -  & -& - & - & - & - \\
& \multicolumn{1}{l}{Simple Incorporation(644M)} & 41.79 & 49.67 & 52.25 & 47.77 & 50.30 & 46.88 & 76.64 & 74.17 & 54.93 \\
&\multicolumn{1}{l}{\textbf{UMLF(644M)}} & $\textbf{55.22}$ & \underline{$63.25$} & $\textbf{62.74}$ & $\textbf{59.23}$  & 59.32 & $\underline{55.62}$ & $\underline{85.86}$ & $\underline{84.23}$ & \textbf{65.68} \\
&\multicolumn{1}{l}{\qquad-only STAC} &\underline{54.29} &61.24  & 60.84 & 56.38 & \underline{60.41} & 55.20 & 76.00 & 73.82 & 62.27 \\
&\multicolumn{1}{l}{\qquad-only Molweni} &53.79 & \textbf{\textit{63.93}}  & 56.90 & 52.31 & 52.68 & 47.29 & 74.20 & 72.33 & 59.17 \\
&\multicolumn{1}{l}{\qquad-only Doc2Dial} &54.21 &61.67  & \underline{62.32} & \underline{58.41} & 55.84 & 52.63 & 81.53 & 79.36 & 63.25 \\
&\multicolumn{1}{l}{\qquad-only TIAGE} &53.46 &61.13  & 54.35 & 49.43 & \textbf{60.81} & \textbf{57.91} & 77.20 & 75.13 & 61.18 \\
 &\multicolumn{1}{l}{\qquad-only DialSeg711} &52.03 &61.87  & 56.79 & 51.94 & 52.41 & 49.45 & \textbf{92.10} & \textbf{90.96} & 63.44 \\
\midrule
\midrule
\multirow{4}{*}{\textit{{\begin{tabular}[c]{@{}c@{}}Billions-\\ Parameter LLMs\end{tabular}}}} 
& \multicolumn{1}{l}{Qwen2-1.5B-Instruct} & 57.16 & 61.23 & 57.32 & 24.49 & 47.95 & 15.12 & 49.17 & 27.13 & 42.45 \\
& \multicolumn{1}{l}{\qquad+parsing module} & 52.17 & 53.80 & 52.38 & 49.52 & 51.47 & 45.98 & 49.61 & 45.45 & 50.05 \\
& \multicolumn{1}{l}{\qquad+UMLF} & 56.10 & 60.34 & 58.27 & 56.62 & 56.40 & 51.28 & 55.02 & 49.03 & 55.38 \\
& \multicolumn{1}{l}{Llama3-8B-Instruct} & 57.04 & 60.59 & 60.16 & 49.07 & 50.71 & 41.19 & 59.52 & 33.82 & 51.51 \\
& \multicolumn{1}{l}{\qquad+parsing module} & 56.82 & 60.26 & 52.84 & 48.36 & 50.39 & 47.67 & 59.02 & 56.65 & 54.00 \\
& \multicolumn{1}{l}{\qquad+UMLF} & 59.90 & 62.75 & 54.43 & 47.36 & 55.50 & 52.44 & 67.89 & 65.18 & 58.18 \\
& \multicolumn{1}{l}{GPT-3.5-turbo} & 59.91 & 63.75 & 62.37 & 54.73 & 57.62 & 51.24 & 86.99 & 83.65 & 65.03 \\
& \multicolumn{1}{l}{Llama3-70B-Instruct} & \textbf{\textit{65.74}} & 63.68 & 66.51 & 64.81 & 62.05 & 57.81 & 99.94 & 99.92 & 72.56 \\
& \multicolumn{1}{l}{GPT-4o} & 65.10 & 61.66 & \textbf{\textit{67.84}} & \textbf{\textit{65.54}} & \textbf{\textit{64.07}} & \textbf{\textit{59.26}} & \textbf{\textit{99.97}} & \textbf{\textit{99.96}} & \textbf{\textit{72.93}} \\
\bottomrule
\end{tabular}
}
\end{table*}

\textbf{PLMs Baselines.} We compare our approach against a variety of unsupervised baselines, categorized into two main areas: unsupervised topic segmentation and unsupervised discourse parsing. For unsupervised topic segmentation, we consider three prominent models: \textbf{TeT + CLS}~\cite{xu2021topic}, which uses BERT to compute semantic similarity between utterances; \textbf{CSM}~\cite{xing2021improving}, which evaluates topic similarity through the coherence of utterance pairs; and \textbf{DialSTART}~\cite{gao2023unsupervised}, which identifies topic boundaries by leveraging topic-aware utterance representations and assessing coherence and consistency. For unsupervised discourse parsing, given the limited existing work on dialogue, we use the latest approach \textbf{PLM4DiscStruct}~\cite{gao2023unsupervised}, which leverages attention matrices from pre-trained language models for discourse parsing, as our baseline. Additionally, we take a \textbf{Simple Incorporation} method as an ablation baseline. It directly merges rhetorical and topic matrices through simple addition, providing a fair comparison under the same scale. To further investigate the performance of our method on each dataset, we also conducted five sets of comparative experiments using a single dataset as the training set, as shown in Table \ref{main} -only rows.

\textbf{LLMs Baselines.} The baseline for the LLMs is divided into two parts. The first part follows Fan et al.~\cite{fan-etal-2024-uncovering-potential}, who directly prompted the LLMs to infer the structure in the format of text. We utilize prior work's~\cite{fan-etal-2024-uncovering-potential} prompts for the LLM to generate two structures. Experiments were conducted on \textbf{Qwen2-1.5B-Instruct, Llama3-8B-Instruct/70B-Instruct, GPT-3.5-turbo, GPT-4o}. The specific prompt formats are detailed in Appendix \ref{prompt}. Moreover, to evaluate the effectiveness of our method on LLMs, we applied the whole framework in Fig. \ref{main} but without the Mutual Learning Module on LLMs (\textbf{Qwen2-1.5B-Instruct and Llama3-8B-Instruct}) to establish a baseline. In detail, we apply prompt engineering~\cite{le2021many} in the same manner as the first baseline on the LLMs to obtain representations and matrices for the two structures. These matrices are then input into the parsing module to derive the corresponding structures. By extracting the final token of each utterance, we obtain representations containing rhetorical and topical information. The computations between these representations yield matrices similar to $A^{top}$ and $A^{rhe}$ in Fig. \ref{main}. Finally, the two matrices are directly fed into the parsing module to generate the corresponding structures. A detailed discussion of this second LLM baseline, as well as the application of UMLF on LLMs, will be provided in subsequent sections.

\textbf{UMLF Application on LLMs.}\label{UMLF_LLMs} To obtain matrices containing rhetorical and topical information for each utterance, which are used for consistency computation and structural parsing, we choose prompt-enigneering~\cite{le2021many} as the unsupervised learning method for LLMs (Llama3-8B-Instruct), replacing the two encoders in the PLMs method. We leverage the prompts from Fan et al.~\cite{fan-etal-2024-uncovering-potential} to guide the Llama3-8B-Instruct in generating outputs with rhetorical and topic information, while these prompts are inserted at the beginning of each dialogue. In our approach, we propose using the final token of each sentence in the model's output as the representation. The rhetorical and topic matrices are then derived through computations between these representations. These matrices are directly fed into the parsing module to generate the corresponding structures, resulting in a baseline without mutual learning, as shown in the configuration of \textbf{LLM + parsing module} in Table \ref{main}. For comparison, we feed these matrices into the mutual learning module, as illustrated in Fig. \ref{main}, and perform backpropagation using the loss calculated by Equation \ref{equation5}, thereby training the entire model. The common matrix obtained after training is then passed through the parsing module to derive the rhetorical and topic structures, representing the application of UMLF on large models, as shown in the configuration of \textbf{LLM + UMLF} in Table \ref{main}.

\textbf{Implementation Details.} \label{implementation}
About PLM-based implementations, following the previous work~\cite{li2023discourse} and \cite{gao2023unsupervised}, for the rhetorical pre-trained model, we adopt the Bart-Large-CNN model\footnote{\url{https://github.com/chuyuanli/PLM4DiscStruct/blob/main/main.ipynb}}~\cite{lewis-etal-2020-bart}, which is fine-tuned on the summarization task of CNN-Dailymail. In terms of the topic pre-trained model selection, we are consistent with the original paper\footnote{\url{https://github.com/AlibabaResearch/DAMO-ConvAI/tree/main/dial-start}}, which used a topic-encoder and a coherence-encoder to separately obtain the consistency and coherence scores. We employed sup-simcse-bert-base-uncased~\cite{gao2021simcse} and NSPBert~\cite{devlin-etal-2019-bert} for topic-encoder and coherence-encoder, respectively. 

Given that we expanded the topic relationship from between adjacent sentences to across non-adjacent sentences, our computational resources have to be extended from n to n square accordingly. With this extension, to avoid out-of-memory issues, we truncate all the training and validation datasets (prior to sampling in Datasets), ensuring that each dialogue has fewer than 18 turns. Nonetheless, the test set continues to test dialogues of all lengths.

It is important to note that for the UMLF application on LLMs, our Qwen2-1.5B-Instruct and Llama3-8B-Instruct models utilize LoRA fine-tuning. During the fine-tuning process, we experimented with three configurations of LoRA settings: the LoRA rank (r) = 8, the LoRA alpha = 16; r = 16, alpha = 32; r = 32, alpha = 64. Among these, the r = 16, alpha = 32 configuration demonstrated the best performance and was selected for the final results. The LoRA dropout ratio is 0.1. Furthermore, LoRA target modules are limited to the attention blocks, specifically the $q_{proj}$ and $v_{proj}$ layers. Notably, the number of trainable parameters for LoRA is 8.6M on Qwen-1.5B-Instruct and 104M on Llama3-8B-Instruct.

In all experiments, throughout the training process on an A-800 GPU, Adam is utilized with a learning rate of 3e-6 and penalty parameters are both set to 1e-3. Early stopping is aligned within three epochs of model training to prevent overfitting and optimize computational efficiency. For reproducibility, we set the seed to 42 in all the experiments.

\subsection{Main Results}

Table \ref{main} presents the performance of our model (UMLF) compared to the baselines across five datasets. Unlike simple incorporation methods, which did not yield any additional performance improvements, UMLF achieves superior performance on all tasks due to the impact of our mutual learning module. Specifically, UMLF outperforms all baselines with significant improvements of 12.32\% and 14.33\% on the STAC and Molweni datasets. UMLF also demonstrates notable improvements in topic segmentation: 2.79\% (1-$P_k$) and 3.11\% (1-WD) on Doc2Dial, 3.66\% (1-$P_k$) and 5.38\% (1-WD) on TIAGE, and 3.72\% (1-$P_k$) and 4.03\% (1-WD) on DialSeg711. Notably, our method achieves over 90 in performance on the relatively simple topic segmentation dataset DialSeg711, which concatenates dialogues from multiple datasets for topic segmentation. This significant performance demonstrates the method's practical potential on simple datasets and their downstream tasks.

To verify the effectiveness of our approach on different types of data (rhetorical, topical), we trained five variant models on individual datasets (-only in Table \ref{main}). For example, -only STAC indicates that our UMLF model is trained solely on the STAC dataset for mutual learning. Compared with millions-parameter PLMs, these models achieved the second-best performance (as indicated by the underline) on the STAC and Doc2Dial datasets and achieved the best performance (as indicated by the bold) on the Molweni, TIAGE, and DialSeg711 datasets. This suggests that our method effectively enhances the modeling of simple discourse relations across various datasets. However, on more challenging datasets—such as STAC, which features a higher number of relations per dialogue, and Doc2Dial, which involves more frequent topic shifts (as shown in Table \ref{tab:dataset_splits_by_type})—such improvements are relatively smaller.

Moreover, our million-parameter UMLF model, without additional annotation, is already comparable to large language models (LLMs) that are significantly larger in scale. UMLF outperforms Qwen2-1.5B-Instruct and Llama3-8B-Instruct across all metrics except for STAC and achieves an almost identical mean performance to GPT-3.5-turbo (65.68 vs. 65.03), as shown in Table \ref{main}. Even when compared to more advanced and larger models like Llama3-70B-Instruct and GPT-4, our model achieves 90\% of their performance (65.68 vs. 72.56/72.93). These results indicate that UMLF is both efficient and effective for practical applications. 

It is worth noting that smaller LLMs with weaker capabilities, such as Qwen2-1.5B-Instruct and Llama3-8B-Instruct, perform significantly worse on the 1-WD metric compared to the 1-$P_k$ metric. This discrepancy occurs because they tend to segment a single sentence into an entire topic when directly generating topic-structured texts, making them more prone to errors when calculating window boundaries in the 1-WD metric. In contrast, decoding algorithms like TextTiling can constrain the number of boundaries, leading to more rigorous topic segmentation.

In summary, the mutual learning process between rhetorical and topical structures, facilitated by the unified representation and linguistic relationships, helps mitigate error accumulation. This allows our method to surpass corresponding PLM baselines and achieve performance comparable to larger-scale LLMs. Further analysis for the effectiveness of two hypotheses can be found in Section \ref{analysis}.

\subsection{UMLF Application Results on LLMs}\label{UMLF_LLM}

Table \ref{main} demonstrates that our method is effective on large language models, showing varying degrees of improvement on both 1.5B and 8B models. For Qwen2-1.5B-Instruct, the STAC dataset shows a 3.93\% increase, the Molweni dataset improves by 6.54\%, and the Doc2Dial dataset exhibits gains of 5.89\% and 7.10\% on the 1-$P_k$ and 1-WD metrics, respectively. Similarly, for the TIAGE dataset, the improvements are 4.93\% and 5.30\% for these metrics, while DialSeg711 shows gains of 5.41\% and 3.58\%. For Llama3-8B-Instruct, the STAC dataset improves by 3.08\%, Molweni by 2.49\%, and TIAGE by 5.11\% and 4.77\% on the 1-$P_k$ and 1-WD metrics, respectively. DialSeg711 shows even more pronounced gains, with improvements of 8.87\% and 8.53\%, while Doc2Dial exhibits a 1.59\% increase on 1-$P_k$, despite a slight 1.00\% decrease on 1-WD. These results collectively confirm the robustness of our method across different model sizes.

We also observe that Llama3-8B performs less effectively on the Doc2Dial dataset, particularly on the 1-WD metric. This can be attributed to the dataset's characteristics, such as the limited number of utterances per topic, as highlighted in Shif. in Table \ref{tab:dataset_splits_by_type}. This property tends to bias the model toward predicting each utterance as a separate topic, making it more challenging to effectively group related utterances.

Overall, these results demonstrate the effectiveness of our method on large language models and validate its generalizability across different model sizes, showcasing its potential to enhance topic segmentation performance.

\section{Analysis}\label{analysis}

To better validate our hypotheses, we conduct a further analysis of the PLM-based results. In Section \ref{rhe_sup}, we first examine the contribution of rhetorical structure to topic structure by analyzing the performance of our model on topic datasets to validate Hypothesis 1. In Section \ref{topic_sup}, we then analyze the impact of topic structure on rhetorical structure by evaluating the performance of our model on rhetorical datasets to validate Hypothesis 2. To validate the effectiveness of mutual learning in our method, in Section \ref{mutual}, we analyze the relative changes between topic and rhetorical structures throughout the entire training phase. Two case studies from STAC and TIAGE are shown in Section \ref{case_study}.

\subsection{The Contribution of Rhetorical Structure to Topic Structure}\label{rhe_sup}

To validate Hypothesis 1 (Local Discourse Coupling), which posits that local rhetorical information contributes to defining topic boundaries, we introduce a new variable termed \textbf{Local Rhetorical Intensity}. This variable represents the density of rhetorical relations within a dialogue. We hypothesize that rhetorical intensity increases with the strength of rhetorical relationships and decreases as the distance between related utterances grows. Therefore, the rhetorical intensity of each dialogue is calculated by summing all the relationship values in the rhetoric-assisted topic matrix \(A_{m}^{top_{rhe}}\), normalized by their corresponding distances.

\begin{figure}[htbp]
\centering
\includegraphics[width=0.8\linewidth]{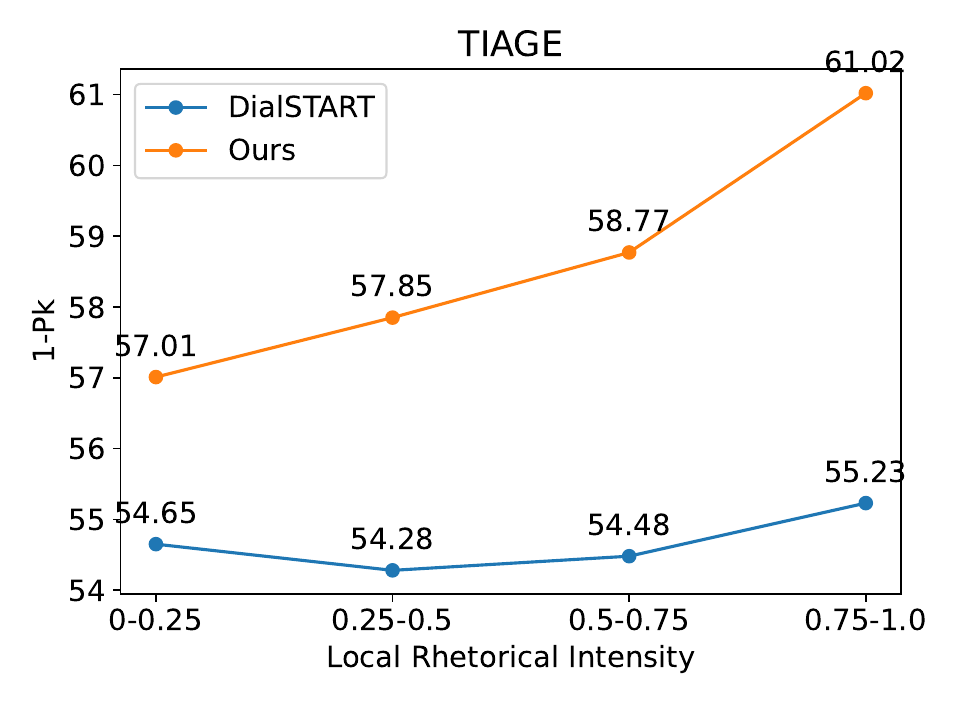} 
\caption{Rhetoric-assisted topic analysis using Local Rhetorical Intensity on the TIAGE dataset. Rhetorical Intensity denotes the degree of clustering in the rhetorical matrix based on the numerical values of its cells. A higher rhetorical intensity facilitates clearer segmentation of the rhetorical matrix into distinct blocks.} 
\label{fig:top_ana}
\end{figure}

To explore the impact of different levels of local rhetorical intensity on topic structure, we apply the following procedure on the TIAGE dataset. First, we calculate the local rhetorical intensity for each test sample’s rhetoric-assisted topic matrix \(A_{m}^{top_{rhe}}\). Based on this intensity, we divide the test set into four intervals: 0-0.25, 0.25-0.5, 0.5-0.75, and 0.75-1.0. We then calculate the topic segmentation results for samples within each interval and perform the same analysis on the results from the DialSTART method for comparison. The results are depicted in Fig.~\ref{fig:top_ana}.

Our analysis reveals that our method consistently outperforms the baseline DialSTART, which does not consider rhetorical relationships in topic segmentation, across all levels of local rhetorical intensity. This indicates that our mutual learning approach effectively leverages local rhetorical information to enhance topic structure identification. Furthermore, as local rhetorical intensity increases, the 1-$P_k$ result for DialSTART remains stable, while our method shows continuous improvement. This suggests that our approach can more effectively utilize local rhetorical intensity—specifically, the rhetorical relationships between an utterance and its surrounding context—to improve the identification of topic boundaries.

\subsection{The Contribution of Topic Structure to Rhetorical Structure}\label{topic_sup}

To validate Hypothesis 2, which posits that topic structures can probabilistically constrain rhetorical structures, we examine the impact of mutual learning on discourse parsing with varying arc distances. Also, we analyze the results across different dialogue lengths, as we hypothesize that dialogues with more turns may be more affected by complex relationships.

\begin{figure}[htbp]
\centering
\subfloat[STAC: F1 and arcs distance]{\includegraphics[width=0.5\linewidth]{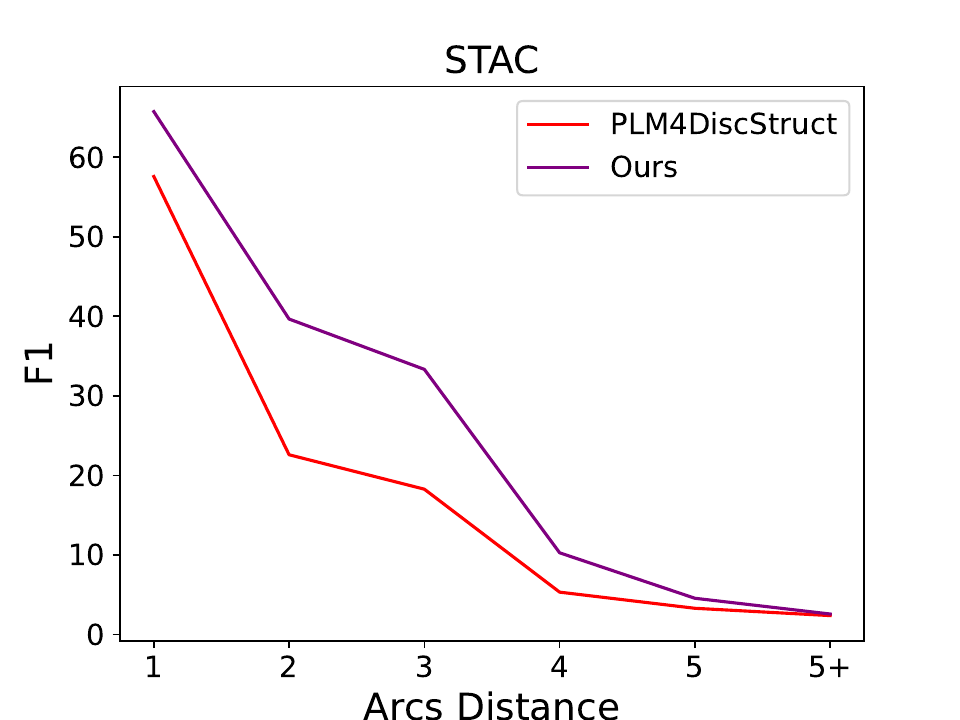}}
\hfill
\subfloat[Molweni: F1 and arcs distance]{\includegraphics[width=0.5\linewidth]{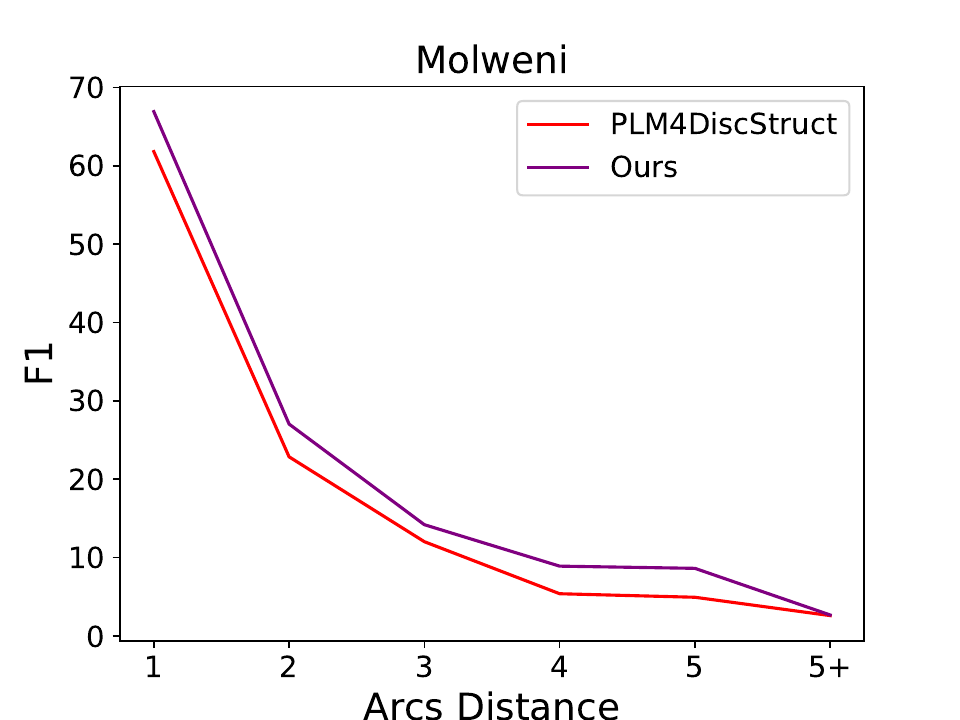}}\\
\subfloat[STAC: F1 and dialogue turns]{\includegraphics[width=0.5\linewidth]{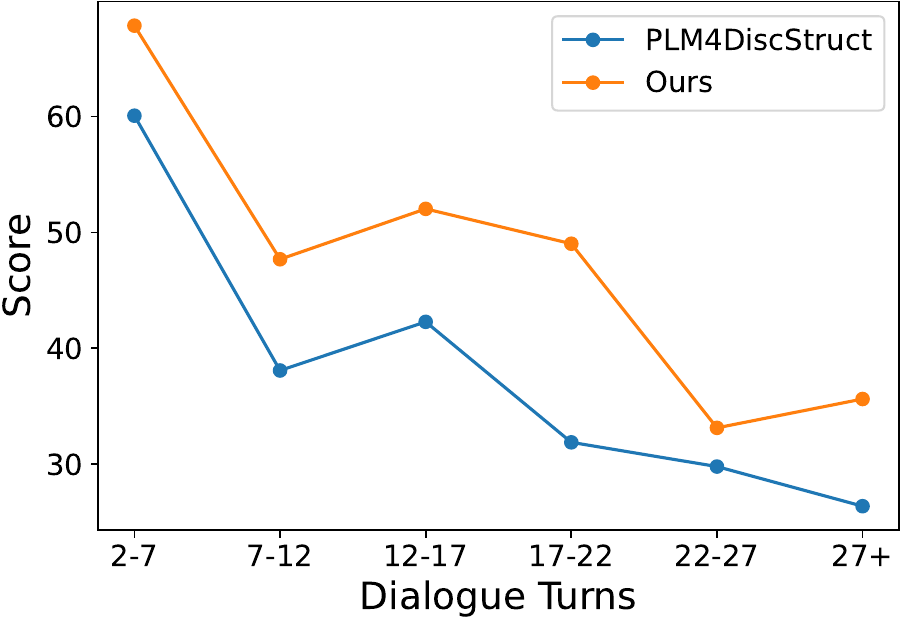}}
\hfill
\subfloat[Molweni: F1 and dialogue turns]{\includegraphics[width=0.5\linewidth]{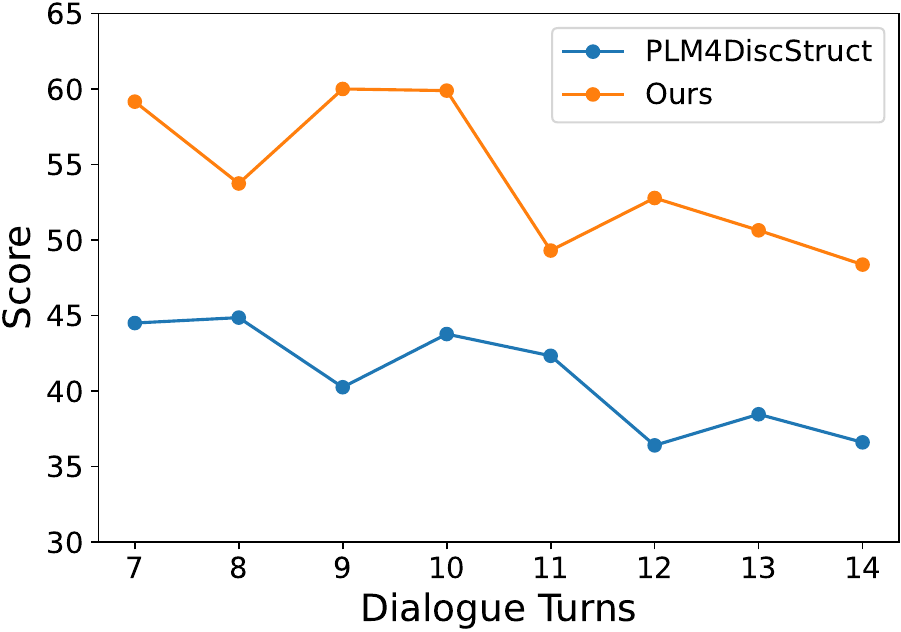}}
\caption{Arc distance analysis with F1, as well as dialogue turns analysis with F1 in STAC and Molweni.}
\label{fig:four_figures}
\end{figure}

\textbf{Arc Distances.} For the STAC and Molweni datasets, we first categorize the lengths of rhetorical relationship arcs into five intervals (1 to 5) and a sixth interval for lengths greater than 5. The results, depicted in Fig.~\ref{fig:four_figures} (a) and (b), demonstrate that for arc distances less than 5, our method significantly outperforms the baseline without mutual learning in terms of F1 scores. We attribute this improvement to the enhancement of rhetorical relationships by the topic structure during the mutual learning process, particularly among utterances spanning 3 to 5 turns. This aligns with the topic distribution in the dataset shown in Table \ref{tab:dataset_splits_by_type}, where the average topic length across the three topic datasets is between 3 and 5, suggesting that rhetorical relations within the same topic have been strengthened.

\textbf{Dialogue Lengths.} Longer dialogues typically pose greater challenges in capturing rhetorical relationships. To examine how our method performs across different dialogue lengths, we divide dialogues in the STAC dataset into six intervals following previous work~\cite{li2023discourse}: five intervals covering spans of 5 turns each (from 2 to 27 turns), and a final interval for dialogues exceeding 27 turns. For the Molweni dataset, each turn count is treated as an individual interval. As shown in Fig.~\ref{fig:four_figures} (c) and (d), our mutual learning approach consistently achieves higher F1 scores compared to the baseline across all intervals. This improvement highlights our method's ability to leverage the global information provided by topic structures, which not only enhances rhetorical structuring in short dialogues but also leads to improvements in long dialogues through the macro-level control of global topic information.

These findings underscore the effectiveness of integrating topic structure into rhetorical parsing, particularly in dialogues where topic segments align closely with rhetorical arcs.

\subsection{The Effectiveness of Mutual Learning}\label{mutual}

\begin{figure}[htbp]
    \centering
    \subfloat[Beginning epoch: topic, rhetorical, and common structures.]{
        \includegraphics[width=\linewidth]{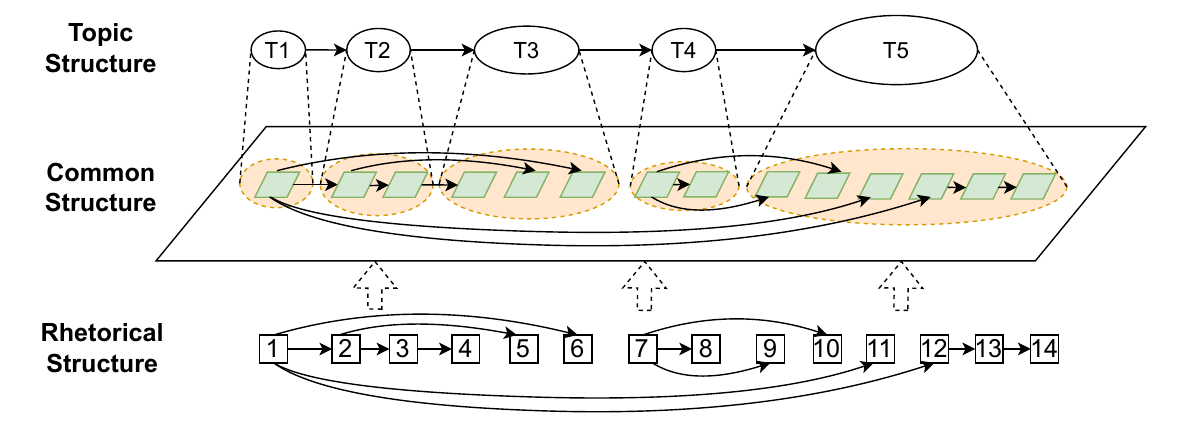}
        \label{fig:mutual_ana1}
    }\\
    \subfloat[Middle epoch: topic, rhetorical, and common structures.]{
        \includegraphics[width=\linewidth]{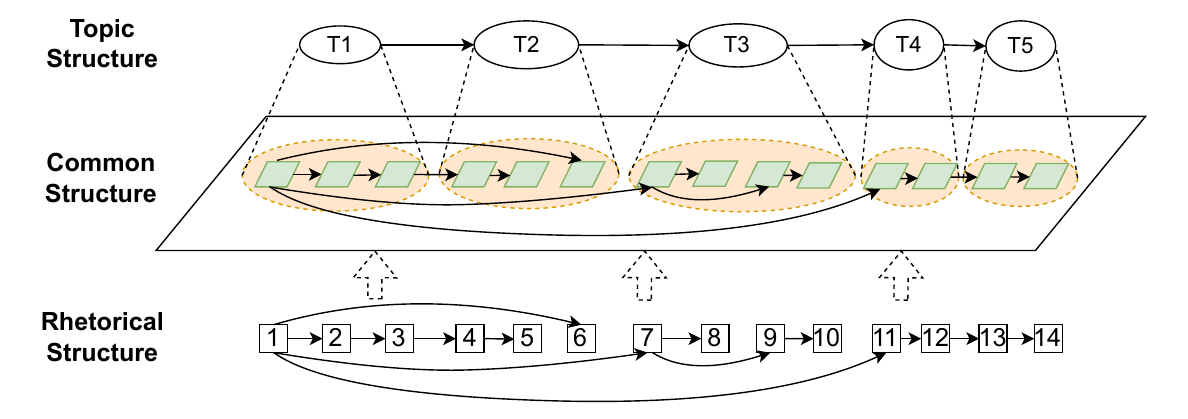}
        \label{fig:mutual_ana2}
    }\\
    \subfloat[End epoch: topic, rhetorical, and common structures.]{
        \includegraphics[width=\linewidth]{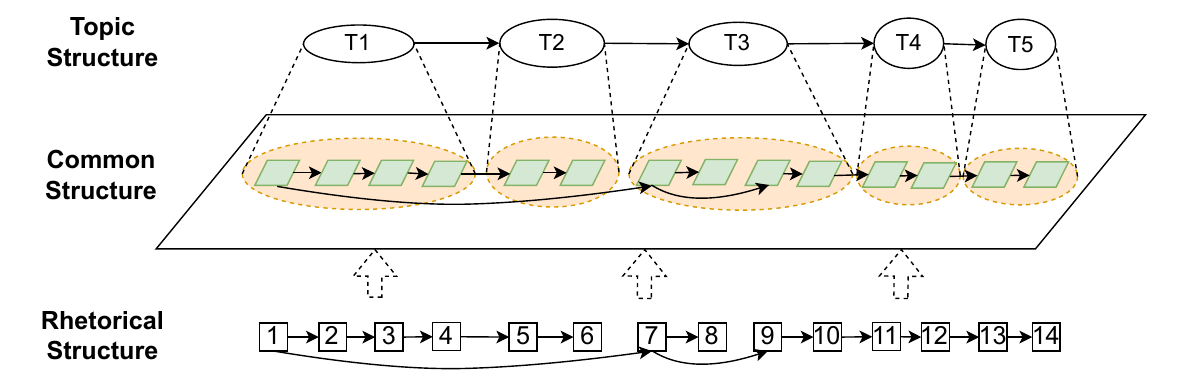}
        \label{fig:mutual_ana3}
    }
    \caption{Evolution of the topic, rhetorical, and common structures during the training process.}
    \label{fig:mutual_ana}
\end{figure}

To verify that our improvements in both rhetorical and topic structures result from reducing the distance between the two composite structures and the common structure, we extract the topic and rhetorical structures from the composite structures at the beginning, middle, and end of the training process and compare them with each other. We analyze an example from the TIAGE dataset in detail, as shown in Fig.~\ref{fig:mutual_ana}. The corresponding utterances are displayed in Fig.~\ref{fig:cs2}.

It can be seen that the alignment between topic and rhetorical structures within the common structure becomes more pronounced as training progresses. This is evidenced by a decrease in rhetorical relationships outside topic boundaries and an increased tendency for utterances with closely linked rhetorical relationships to coalesce into a single topic. These observations are consistent with our two hypotheses: topic structures enhance rhetorical relationships within topic boundaries while diminishing those outside, and rhetorical relationships, in turn, influence topic boundaries based on their local distribution. Furthermore, the patterns observed during training confirm that our alignment operation effectively facilitates mutual learning between the two structures.

\subsection{Case Study}\label{case_study}

\begin{figure*}[htbp]
    \centering
    \subfloat[Gold Rhetorical Structure.]{
        \includegraphics[width=0.3\linewidth]{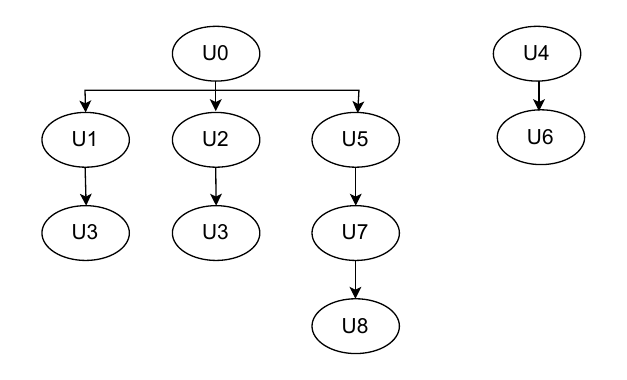}
    }
    \hfill
    \subfloat[Ours Rhetorical Structure.]{
        \includegraphics[width=0.2\linewidth]{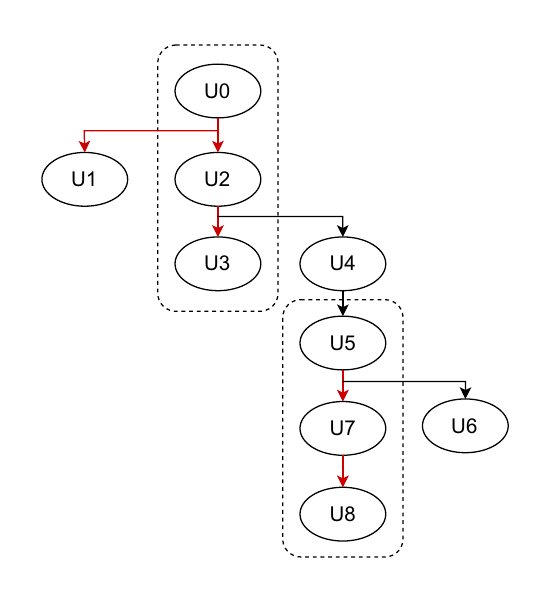}
    }
    \hfill
    \subfloat[PLM4DiscStruct Rhetorical Structure.]{
        \includegraphics[width=0.3\linewidth]{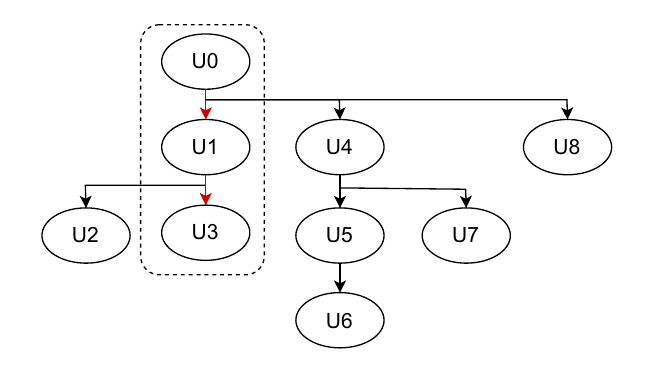}
        \label{fig:3}
    }
    \caption{Case study of discourse parsing from STAC, dialogue\_id: s2-league4-game2. The concrete utterances are shown in Fig.~\ref{fig:case2}. The order from left to right represents the rhetorical structures generated by the gold standard, our model, and PLM4DiscStruct. The red lines indicate correct relations.}
    \label{fig:cs1}
\end{figure*}

\begin{figure*}[htbp]
    \centering
    \includegraphics[width=0.6\textwidth]{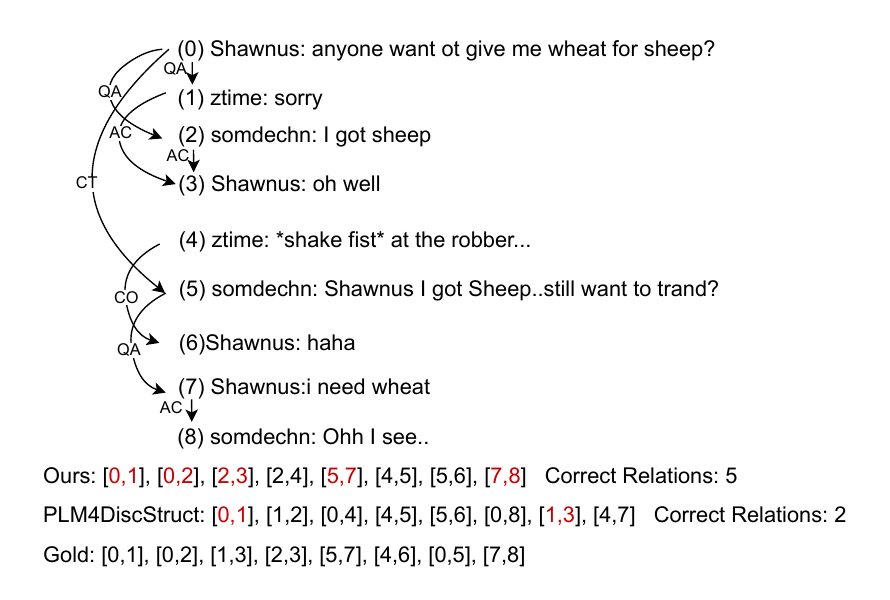}
    \caption{Details of dialogue utterances in discourse parsing from STAC, dialogue\_id: s2-league4-game2. Regarding the rhetorical relation name, QA means Question Answer Pair, AC means Acknowledgement; CO means Comment, and CT means Continuation. The upper part shows the dialogue texts, the gold rhetorical structure, and the relationship names. The lower part presents the results from our method, the baseline (PLM4DiscStruct), and the gold standard, where [0,1] represents a rhetorical relation between utterance 0 and utterance 1. The "Correct Relations" metric refers to correct relationship numbers.}
    \label{fig:case2}
\end{figure*}

\begin{figure*}
    \centering
    \includegraphics[width=0.9\textwidth]{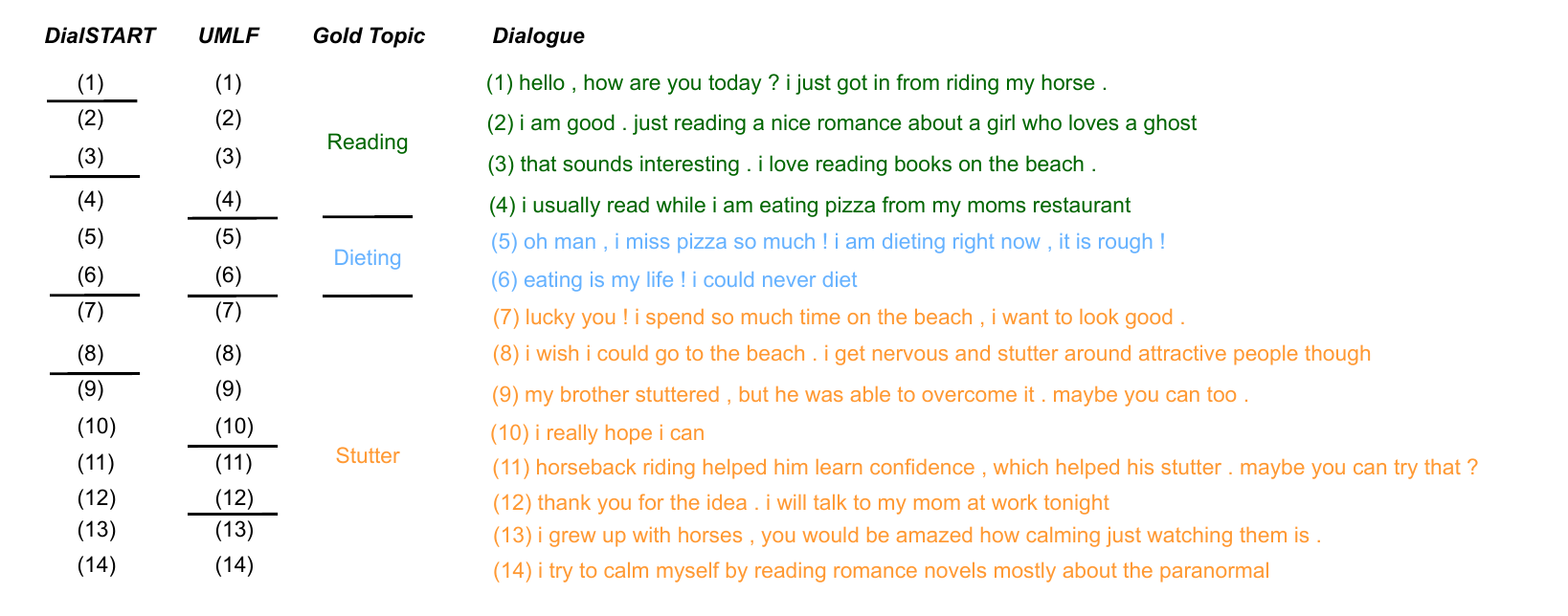}
    \caption{The case study of topic segmentation from TIAGE, dialogue\_id: 78. '--' represents the topic boundary.}
    \label{fig:cs2}
\end{figure*}

We conduct two case studies from STAC and TIAGE to further analyze the effectiveness of our approach. In Fig.~\ref{fig:cs1}, we present the rhetorical tree structures of the gold standard, our method, and BART from left to right, replacing utterances in the dialogue with $u_i$. The concrete utterances are provided in Fig.~\ref{fig:case2}. In Fig.~\ref{fig:cs2}, we display the topic segmentation of the gold standard, our method, and DialSTART from top to bottom, representing utterances in the dialogue with $t_i$. Different colored blocks indicate distinct topics.

For the discourse parsing case from STAC, as shown in Fig. \ref{fig:case2}, our method identifies three more correct relations than the baseline (BART) and successfully detects gold-standard relations on [5, 7], [0, 2], [2, 3], and [7, 8], all of which have arc distances of less than 3. In Fig. \ref{fig:cs1}, our model produces one additional correct rhetorical structure block (a path composed of multiple correct rhetorical relations), demonstrating that topic-assisted rhetoric effectively strengthens rhetorical relationships within a topic.

For the topic segmentation case from TIAGE, our method outperforms the baseline in both $P_k$ and WD metrics. Compared to the baseline (DialSTART), our method successfully identifies the boundary between \textit{Reading} and \textit{Dieting}, leading to improved topic consistency between the boundaries. These results further validate Hypothesis 2, confirming that topic segmentation benefits from strengthened rhetorical relationships.

\begin{table*}[htbp]
\centering
\caption{Prompting LLM to generate a rhetorical structure.}
\label{tab:rhe_prompt}
\begin{tabular}{|c|p{15cm}|}
\hline
\textbf{Prompt} & According to the Segmented Discourse Rhetorical Theory, the rhetorical structure of a dialogue can be represented by a directed acyclic graph, where nodes are utterances and edges are the following 16 relations: \\
& \{ 'Comment':'', "Clarification\_question":"", "Elaboration":"", "Acknowledgement":"", "Explanation":"", "Conditional":"", "Question-answer\ pairs":"", "Alternation":"", "Question-Elaboration":"", "Result":"", "Background":"", "Narration":"", "Correction":"", "Parallel":"", "Contrast":"", "Continuation":"" \} \\
         & Please annotate the rhetorical structure of the following dialogue and represent it in the form of [index1, index2, 'relation'], where index1 and index2 are the index of two utterances, and the 'relation' is one of the above relations to connect the two utterances. \\
\hline
\end{tabular}
\end{table*}

\begin{table*}[htbp]
\centering
\caption{Prompting LLM to generate a topic structure.}
\label{tab:top_prompt}
\begin{tabular}{|c|p{15cm}|}
\hline
\textbf{Prompt} & Please identify several topic boundaries for the following dialogue, and each topic consists of several consecutive utterances. please output in the form of {'topic i':[], ... ,'topic j':[]}, where the elements in the list are the index of the consecutive utterances within the topic, and output even if there is only one topic.\\
\hline
\end{tabular}
\end{table*}

\section{Conclusion and Future Work}
In this paper, we introduce a unified framework that integrates rhetorical and topic structures, drawing upon linguistic theory to formulate two hypotheses regarding their mutual influence. To model these structures jointly, we propose an unsupervised mutual learning framework (UMLF), which employs two aggregators to capture rhetorical and topic information from both local and global perspectives. Additionally, an alignment mechanism is utilized to generate a unified structure that simultaneously encompasses the characteristics of both rhetorical and topic structures.

Our experimental results across five diverse datasets demonstrate that UMLF outperforms existing PLM-based baselines, yielding an average improvement of approximately 12\% on rhetorical datasets and 4\% on topic datasets, and achieves competitive performance of LLM-based models, such as GPT-3.5-turbo and GPT-4o. Notably, experiments with LLMs (Qwen2-1.5B-Instruct and Llama3-8B-Instruct) further confirm the effectiveness and generality of our approach.

In-depth analyses validate both of our hypotheses, illustrating that UMLF successfully aligns rhetorical and topic structures across various dialogue contexts. This establishes the robustness of our framework. Moving forward, we plan to apply UMLF to real-world dialogue systems, especially in fields requiring advanced discourse understanding, such as medical and customer service domains.

\section*{Limitations}\label{limitations}

One limitation of our study is that the current rhetorical and topic datasets are derived from dialogues in relatively simple, single-domain settings, leaving a gap in evaluating our approach to more complex, everyday conversations that simultaneously involve rhetorical and topic relationships. Furthermore, based on the unsupervised setting, we do not conduct robust experimental evidence and validation regarding the mutual influence of rhetorical and topic structures in supervised scenarios, which could be explored in further work. Lastly, we plan to further extend our research to downstream tasks, such as dialogue generation, to better explore the impact and benefits of discourse structures on dialogue systems.

\section*{Acknowledgement}
This research is supported by the project of Shenzhen Science and Technology Research Fund (Fundamental Research Key Project Grant No. JCYJ20220818103001002), Shenzhen Science and Technology Program (Grant No. ZDSYS20230626091302006), Key Project of Shenzhen Higher Education Stability Support Program (Grant No. 2024SC0009), SRIBD Innovation Fund (Grant No. K00120240006), and Shenzhen Science and Technology Program (Grant No. RCBS20231211090538066), and European Commission through Project ASTOUND (101071191 —HORIZON-EIC-2021-PATHFINDERCHALLENGES-01), and project BEWORD (PID2021-126061OB-C43) funded by MCIN/AEI/10.13039/501100011033 and, as appropriate, by “ERDF A way of making Europe”, by the “European Union”.

\appendix
\section{LLMs Prompt}\label{prompt}
We test LLMs' performance on five datasets following Fan et al.'s work~\cite{fan-etal-2024-uncovering-potential}. The prompt of rhetorical and topic structure inference are shown in Table \ref{tab:rhe_prompt} and \ref{tab:top_prompt}.

\bibliographystyle{IEEEtran}
\bibliography{IEEEabrv,new}

\end{document}